\documentclass[pmlr]{jmlr}
\usepackage{array}
\usepackage{amsmath} 
\usepackage{amssymb}
\usepackage{bbm}
\usepackage{booktabs}
\usepackage{comment}
\usepackage{graphbox}
\usepackage{hhline}
\usepackage{makecell}
\usepackage{multirow}
\usepackage{natbib}
\usepackage[compact]{titlesec}
\usepackage{wrapfig}
\usepackage{xspace}

\newcommand{\oursfullupper}{Clinically Coherent Reward\xspace}
\newcommand{\oursfull}{\MakeLowercase{\oursfullupper}\xspace}
\newcommand{\oursabbr}{CCR\xspace}
\newcommand{\ourseqn}{\operatorname{\mathrm{\oursabbr}}}

\newcommand{\para}[1]{{\left( #1 \right)}}
\newcommand{\param}[1]{{\left[ #1 \right]}}
\newcommand{\paral}[1]{{\left\{ #1 \right\}}}
\newcommand{\abs}[1]{{\left| #1 \right|}}

\newcommand{\lstm}{\operatorname{\mathrm{LSTM}}}
\newcommand{\relu}{\operatorname{\mathrm{ReLU}}}
\newcommand{\softmax}{\operatorname{\mathrm{softmax}}}
\newcommand{\auto}{\operatorname{\mathrm{NLG}}}

\newcommand{\report}{\mathbf{Z}}
\newcommand{\sent}{\mathbf{z}}
\newcommand{\word}{z}
\newcommand{\voc}{\mathbb{V}}
\newcommand{\expect}{\mathbb{E}}
\newcommand{\loss}{\mathcal{L}}
\newcommand{\real}{\mathbb{R}}
\newcommand{\unit}{\mathbbm{1}}

\newcommand{\absmen}{\mathsf{a}}
\newcommand{\negmen}{\mathsf{n}}
\newcommand{\uncmen}{\mathsf{u}}
\newcommand{\posmen}{\mathsf{p}}

\newcommand{\posdis}{\mathsf{+}}
\newcommand{\negdis}{\mathsf{-}}

\newcommand{\bb}{\mathbf{b}}
\newcommand{\bc}{\mathbf{c}}
\newcommand{\be}{\mathbf{e}}
\newcommand{\bh}{\mathbf{h}}
\newcommand{\bm}{\mathbf{m}}
\newcommand{\bp}{\mathbf{p}}
\newcommand{\bs}{\mathbf{s}}
\newcommand{\bv}{\mathbf{v}}
\newcommand{\bw}{\mathbf{w}}

\newcommand{\bV}{\mathbf{V}}
\newcommand{\bE}{\mathbf{E}}
\newcommand{\bW}{\mathbf{W}}
\newcommand{\balpha}{\boldsymbol\alpha}
\newcommand{\btau}{\boldsymbol\tau}

\newcolumntype{L}[1]{>{\raggedright\let\newline\\\arraybackslash\hspace{0pt}}m{#1}}
\newcolumntype{C}[1]{>{\centering\let\newline\\\arraybackslash\hspace{0pt}}m{#1}}
\newcolumntype{R}[1]{>{\raggedleft\let\newline\\\arraybackslash\hspace{0pt}}m{#1}}

\jmlrproceedings{}{}
\jmlrvolume{}
\jmlryear{}
\jmlrissue{}
\jmlrpages{}
\jmlrworkshop{}

\title[Clinically Accurate Chest X-Ray Report Generation]{Clinically Accurate Chest X-Ray Report Generation}

\author{%
    \Name{Guanxiong Liu}\footnotemark[1]
    \Email{lgx@cs.toronto.edu}
    \addr \\ University of Toronto
    \AND
    \Name{Tzu-Ming Harry Hsu}\footnotemark[1]
    \Email{stmharry@mit.edu}
    \addr \\ Massachusetts Institute of Technology
    \AND
    \Name{Matthew McDermott}
    \Email{mmd@mit.edu}
    \addr \\ Massachusetts Institute of Technology
    \AND
    \Name{Willie Boag}
    \Email{wboag@mit.edu}
    \addr \\ Massachusetts Institute of Technology
    \AND
    \Name{Wei-Hung Weng}
    \Email{ckbjimmy@mit.edu}
    \addr \\ Massachusetts Institute of Technology
    \AND
    \Name{Peter Szolovits}
    \Email{psz@mit.edu}
    \addr \\ Massachusetts Institute of Technology
    \AND
    \Name{Marzyeh Ghassemi}
    \Email{marzyeh@cs.toronto.edu}
    \addr \\ University of Toronto, Vector Institute
}

\editor{}

\begin{document}
\phantom{\thanks{Equal contribution, ordered alphabetically.}}
\maketitle

\begin{abstract}
The automatic generation of radiology reports given medical radiographs has significant potential to operationally and improve clinical patient care. A number of prior works have focused on this problem, employing advanced methods from computer vision and natural language generation to produce readable reports. However, these works often fail to account for the particular nuances of the radiology domain, and, in particular, the critical importance of clinical accuracy in the resulting generated reports. 
In this work, we present a domain-aware automatic chest X-ray radiology report generation system which first predicts what topics will be discussed in the report, then conditionally generates sentences corresponding to these topics. The resulting system is fine-tuned using reinforcement learning, considering both readability and clinical accuracy, as assessed by the proposed Clinically Coherent Reward. We verify this system on two datasets, Open-I and MIMIC-CXR, and demonstrate that our model offers marked improvements on both language generation metrics and CheXpert assessed accuracy over a variety of competitive baselines.
\end{abstract}

\section{Introduction}
A critical task in radiology practice is the generation of a free-text description, or \emph{report}, based on a clinical radiograph (e.g., a chest X-ray). Providing automated support for this task has the potential to ease clinical workflows and improve both the quality and standardization of care. However, this process poses significant technical challenges. Many traditional image captioning approaches are designed to produce far shorter and less complex pieces of text than radiology reports. Further, these approaches do not capitalize on the highly templated nature of radiology reports. Additionally, generic natural language generation (NLG) methods prioritize descriptive accuracy only as a byproduct of readability, whereas providing an accurate clinical description of the radiograph is the \emph{first} priority of the report. Prior works in this domain have partially addressed these issues, but significant gaps remain towards producing high-quality reports with maximal clinical efficacy.

In this work, we take steps to address these gaps through our novel automatic chest X-ray radiology report generation system. Our model hierarchically generates a sequence of unconstrained topics, using each topic to generate a sentence for the final generated report. In this way, we capitalize on the often-templated nature of radiology reports while simultaneously offering the system sufficient freedom to generate diverse, free-form reports. The system is finally tuned via reinforcement learning to optimize readability (via the CIDEr score) as well as clinical accuracy (via the concordance of CheXpert~\citep{irvin2019chexpert} disease state labels between the ground truth and generated reports). We test this system on the MIMIC-CXR~\citep{johnson2019mimic} dataset, which is the largest paired image-report dataset presently available, and demonstrate that our model offers improvements on both NLG evaluation metrics (BLEU~\citep{papineni2002bleu}, CIDEr~\citep{vedantam2015cider}, and ROGUE~\citep{lin2004rouge}) and clinical efficacy metrics (CheXpert concordance) over several compelling baseline models, including a re-implementation of TieNet~\citep{wang2018tienet}, simpler neural baselines, and a retrieval-based baseline.

\paragraph{Clinical Relevance}
This work focuses on generating a clinically useful radiology report from a chest X-ray image. This task has been explored multiple times, but directly transplanting natural language generation techniques onto this task only guarantees the reports to \emph{look real} rather than to \emph{predict right}. A more immediate focus for the report generation task is thus to produce accurate disease profiles to power downstream tasks such as diagnosis and care providing. Our goal is then minding the language fluency while also increasing the clinical efficacy of the generated reports.

\paragraph{Technical Significance}
We employ a hierarchical convolutional-recurrent neural network as the backbone for our proposed method. Reinforcement learning (RL) on a combined objective of both language fluency metrics and the proposed \oursfullupper (\oursabbr) ensures we obtain a quality model on more correctly describing disease states. Our method aims to numerically align the disease labels of our generated report, as produced by a natural language labeler, with the labels from the ground truth reports. The reward function, though non-differentiable, can be optimized through policy gradient learning as promised by RL.

\section{Background \& Related Work}
\subsection{Radiology}

\begin{wrapfigure}{r}{0.43\textwidth}
\vspace{-10px}
\centering
\includegraphics[width=\linewidth]{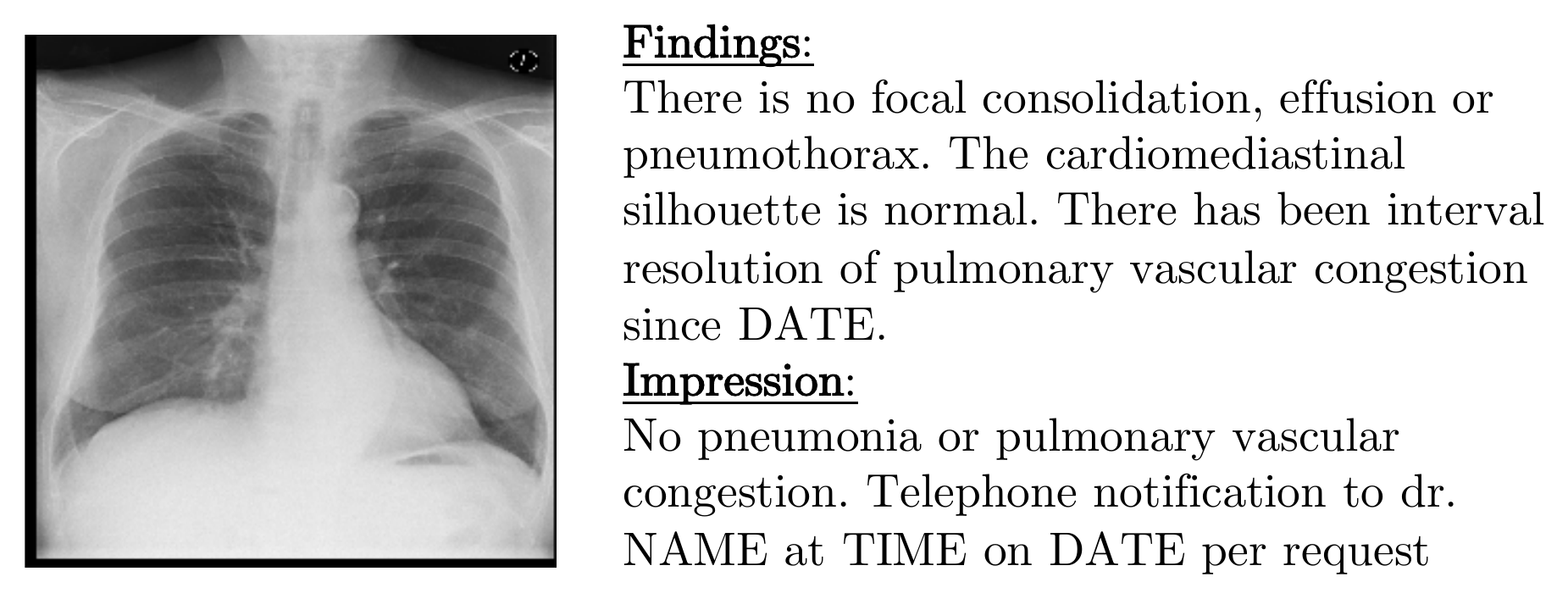}
\caption{\small A chest X-ray and its associated report written by a radiologist.}
\label{fig:illustration}
\vspace{-20pt}
\end{wrapfigure}
\paragraph{Radiology Practice}
Diagnostic radiology is the medical field of creating and evaluating radiological images (radiographs) of patients for diagnostics. 
Radiologists are trained to simultaneously identify various
radiological findings (e.g., diseases),
according to the details of the radiograph and the patient's clinical history, then summarize these findings and their overall impression in reports for clinical communication~\citep{kahn2009toward,schwartz2011improving}. A report typically consists of sections such as \emph{history}, \emph{examination reason}, \emph{findings}, and \emph{impressions}. 
As shown in Figure~\ref{fig:illustration}, the \emph{findings} section contains a sequence of positive, negative, or uncertain mentions of either disease observations or instruments including their detailed location and severity. The \emph{impression} section, by contrast, summarizes diagnoses considering all report sections above and previous studies on the patient. 

Correctly identifying all abnormalities is a challenging task due to high variation, atypical cases, and the information overload inherent to some imaging modalities, such as computerized tomography (CT) scans~\citep{rubin2015lung}.
This presents a strong intervention surface for machine learning techniques to help radiologists correctly identify the critical findings from a radiograph. The canonical way to communicate such findings in current practice would be through the free-text report, which could either be used as a ``draft'' report for the radiologists to extend or be presented to the physician requesting a radiological study directly~\citep{schwartz2011improving}.
%

\paragraph{AI on Radiology Data}
\begin{table}[!t]
\setlength{\tabcolsep}{0.2em}
\centering
\footnotesize
    
\begin{tabular}{L{2.0cm}	L{3.4cm}	L{3.2cm}	R{1.8cm}	R{1.8cm}	R{1.8cm}}
\toprule
\bf Dataset	& \bf Source Institution	& \bf Disease Labeling	& \bf \# Images	& \bf \# Reports	& \bf \# Patients\\ \midrule
Open-I	& Indiana Network for Patient Care	& Expert	& 8,121	& 3,996	& 3,996\\ \hline
Chest-Xray8	& National Institutes of Health	& Automatic \newline (DNorm + MetaMap)	& 108,948	& 0	& 32,717\\ \hline
CheXpert	& Stanford Hospital	& Automatic \newline (CheXpert labeler)	& 224,316	& 0	& 65,240\\ \hline
PadChest	& Hospital Universitario de San Juan	& Expert + Automatic \newline (Neural network)	& 160,868	& 206,222	& 67,625\\ \hline
MIMIC-CXR	& Beth Israel Deacones Medical Center	& Automatic \newline (CheXpert labeler)	& 473,057	& 206,563	& 63,478\\
\bottomrule
\end{tabular}

\caption{\small A description of each available chest X-ray datasets. Open-I~\citep{demner2015preparing}, Chest-XRay8~\citep{wang2017chestx} which utilized DNorm~\citep{leaman2015challenges} and MetaMap~\citep{aronson2010overview}, CheXpert~\citep{irvin2019chexpert}, PadChest~\citep{bustos2019padchest}, and
MIMIC-CXR~\citep{johnson2019mimic}.}
\label{tbl:datasets}
\vspace{-20pt}

\end{table}

In recent years, several chest radiograph datasets, totalling almost a million X-ray images, have been made publicly available. A summary of these datasets is available in Table~\ref{tbl:datasets}. 
%
%
Learning effective computational models through leveraging the information in medical images and free-text reports is an emerging field.
Such a combination of image and textual data help further improve the model performance in both image annotation and automatic report generation~\citep{litjens2017survey}.

\citet{schlegl2015predicting} first proposed a weakly supervised learning approach to utilize semantic descriptions in reports as labels for better classifying the tissue patterns in optical coherence tomography (OCT) imaging.
In the field of radiology, \citet{shin2016learning} proposed a convolutional and recurrent network framework that jointly trained from image and text to annotate disease, anatomy, and severity in the chest X-ray images. 
Similarly, \citet{moradi2018bimodal} jointly processed image and text signals to produce regions of interest over chest X-ray images.
\citet{rubin2018large} trained a convolutional network to predict common thoracic diseases given chest X-ray images.
\citet{shin2015interleaved}, \citet{wang2016unsupervised}, and \citet{wang2017chestx} mined radiological reports to create disease and symptom concepts as labels. 
They first used Latent Dirichlet Allocation (LDA) to identify the topics for clustering, then applied the disease detection tools such as DNorm, MetaMap, and several other Natural Language Processing (NLP) tools for downstream chest X-ray classification using a convolutional neural network. 
They also released the label set along with the image data.

Later on, \citet{wang2018tienet} used the same Chest X-ray dataset to further improve the performance of disease classification and report generation from an image.
For report generation, \citet{jing2017automatic} built a multi-task learning framework, which includes a co-attention mechanism module, and a hierarchical long short term memory (LSTM) module, for radiological image annotation and report paragraph generation.
\citet{li2018hybrid} proposed a reinforcement learning-based Hybrid Retrieval-Generation Reinforced Agent (HRGR-Agent) to learn a report generator that can decide whether to retrieve a template or generate a new sentence.
Alternatively, \citet{gale2018producing} generated interpretable hip fracture X-ray reports by identifying image features and filling text templates.

Finally, \citet{hsu2018unsupervised} trained the radiological image and report joint representation through unsupervised alignment of cross-modal embedding spaces for information retrieval.


\subsection{Language Generation}
Language generation (LG) is a staple of NLP research. LG comes up in the context of neural machine translation, summarization, question answering, image captioning, and more. In all these tasks, the challenges of generating discrete sequences that are realistic, meaningful, and linguistically correct must be met, and the field has devised a number of methods to surmount them. For many years, this was done through ngram-based~\citep{huang1993sphinx} or retrieval-based~\citep{gupta2010survey} approaches.

%

Within the last few years, many have explored the very impressive results of deep learning for text generation. \citet{graves2013generating} outlined best practices for RNN-based sequence generation. The following year, \citet{sutskever2014sequence} introduced the \textit{sequence-to-sequence} paradigm for machine translation and beyond. However, \citet{wiseman2017challenges} demonstrated that while RNN-generated texts are often fluent, they have typically failed to reach human-level quality.

Reinforcement learning recently also come into play due to its capability to optimize for indirect target rewards, even if the targets themselves are often non-differentiable. \citet{li2016deep} used a crafted combination of human heuristics as the reward while \citet{bahdanau2016actor} incorporated language fluency metrics. They were among the first to apply such techniques to neural language generation, but to date, training with log-likelihood maximization~\citep{xie2017neural} has been the main working horse.
Alternatively, \citet{rajeswar2017adversarial} and \citet{fedus2018maskgan} have tried using Generative Adversarial Neural Networks (GANs) for text generation.
However, \citet{caccia2018language} observed problems with training GANs and show that to date, they are unable to beat canonical sequence decoder methods.

\paragraph{Image Captioning}

We will also highlight some specific areas of exploration in image captioning, a specific kind of language generation which is conditioned on an image input. The canonical example of this task is realized in the Microsoft COCO~\citep{coco2014} dataset, which presents a series of images, each annotated with five human-written captions describing the image. The task, then, is to use the image as input to generate a readable, accurate, and linguistically correct caption.

This task has received significant attention with the success of \textit{Show and Tell}~\citep{vinyals2015show} and its followup \textit{Show, Attend, and Tell}~\citep{xu2015show}. 
Due to the nature of the COCO competition, other works quickly emerged showing strong results: \citet{yao2017boosting} used boosting methods, \citet{lu2017knowing} employed adaptive attention, and \citet{rennie2017self} introduced reinforcement learning as a method for fine-tuning generated text.
\citet{devlin2015exploring} performed surprisingly well using a $K$-nearest neighbor method. They observed that since most of the true captions were simple, one-sentence scene descriptions, there was significant redundancy in the dataset.

\subsection{Radiology Report Generation}
Multiple recent works have explored the task of radiology report generation.
\citet{zhang2018learning} used a combination of extractive and abstractive techniques to summarize a radiology report's findings to generate an impression section.
Due to limited text training data, \citet{han2018towards} relied on weak supervision for a Recurrent-GAN and template-based framework for MRI report generation. 
\citet{gale2018producing} uses an RNN to generate template-generated text descriptions of pelvic X-rays.

More comparable to this work, \citet{wang2018tienet} used a CNN-RNN architecture with attention to generate reports that describe chest X-rays based on sequence decoder losses on the generated report.
\citet{li2018hybrid} generated chest X-ray reports using reinforcement learning to tune a hierarchical decoder that chooses (for each sentence) whether to use an existing template or to generate a new sentence, optimizing the language fluency metrics.

\section{Methods}
\begin{figure}[!t]
\centering
\includegraphics[width=\linewidth]{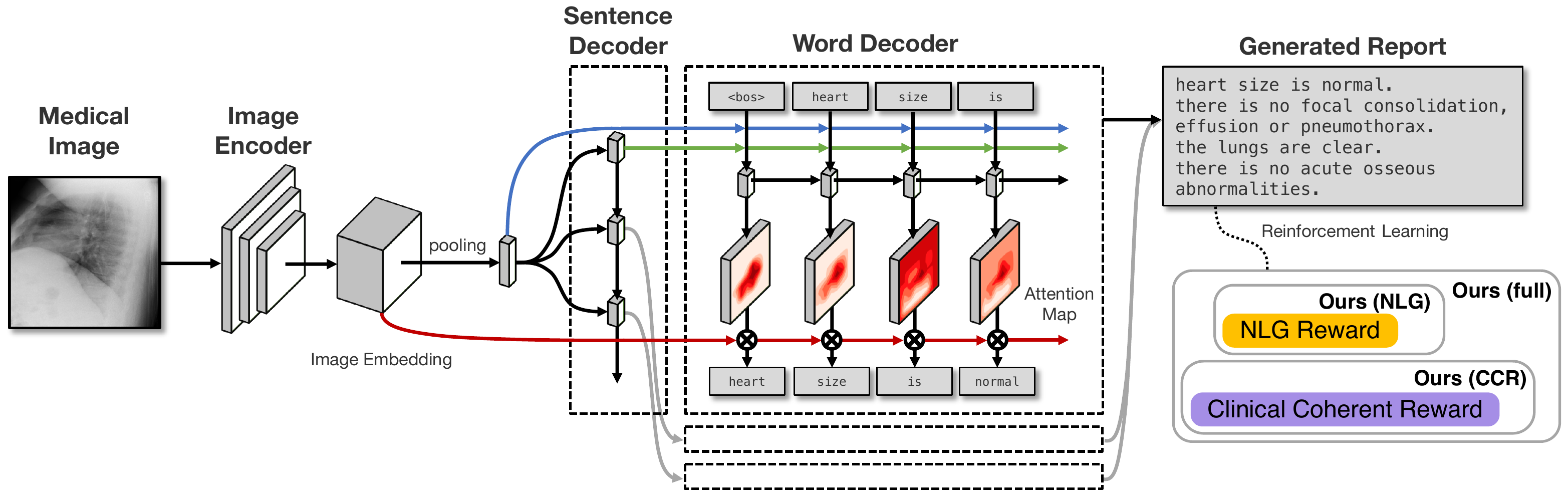}
\caption{\small \textbf{The model for our proposed \emph{\oursfullupper}}. Images are first encoded into image embedding maps, and a sentence decoder takes the pooled embedding to recurrently generate topics for sentences. The word decoder then generates the sequence from the topic with attention on the original images. NLG reward, \oursfull, or combined, can then be applied as the reward for reinforcement policy learning.}
\label{fig:overview}
\vspace{-20pt}
\end{figure}
In this work we opt to focus on generating the \emph{findings} section as it is the most direct annotation from the radiological images. 
First, we introduce the hierarchical generation strategy with a CNN-RNN-RNN architecture, and later we propose novel improvements that render the generated reports more clinically aligned with the true reports. Full implementation details, including layer sizes, training details, etc., are presented in the Appendix, Section~\ref{sec:implementation}.

\subsection{Hierarchical Generation via CNN-RNN-RNN}
As illustrated in Figure~\ref{fig:overview}, we aim to generate a report as a sequence of sentences $\report = \para{\sent_1, \ldots, \sent_M}$, where $M$ is the number of sentences in a report.
Each sentence consists of a sequence of words $\sent_i = \para{\word_{i1}, \ldots, \word_{iN_i}}$ with words from a vocabulary $\word_{ij} \in \voc$, where $N_i$ is the number of words in sentence $i$. 

The image is fed through the \emph{image encoder CNN} to obtain a visual feature map. The feature is then taken by the \emph{sentence decoder RNN} to recurrently generate vectors that represent the topic for each sentence. 
With the visual feature map and the topic vector, a \emph{word decoder RNN} tries to generate a sequence of words and attention maps of the visual features. 
This hierarchical approach is in line with \citet{krause2017hierarchical} where they generate descriptive paragraphs for an image.

\paragraph{Image encoder CNN}
The input image $I$ is passed through a CNN head to obtain the last layer before global pooling, and the feature is then projected to an embedding of dimensionality $d$, which is identical to the word embedding dimension.
The resulting map $\bV = \paral{\bv_k}_{k=1}^K$ of spatial image features will be descriptive features for different spatial locations of an image.
A mean visual feature is obtained by averaging all local visual features $\bar\bv = \frac{1}{K} \sum_k \bv_k$.

\paragraph{Sentence decoder RNN}
Given the mean visual feature $\bar\bv$, we adopt Long-Short Term Memory (LSTM) and model the hidden state as
$
    \bh_i, \bm_i = \lstm\para{\bar\bv; \bh_{i-1}, \bm_{i-1}},
$
where $\bh_{i-1}$ and $\bm_{i-1}$ are the hidden state vector and the memory vector for the previous sentence $\para{i-1}$ respectively. From the hidden state $\bh_i$, we further generate two components, namely the topic vector $\btau_i$ and the stop signal $u_i$ for the sentence, as
$\btau_i = \relu\para{\bW_{\tau}^\top \bh_i + \bb_{\tau}}$
and
$u_i = \sigma \para{\bw_{u}^\top \bh_i + b_{u}}$,
where $\bW$'s and $\bb$'s are trainable parameters, and $\sigma$ is the sigmoid function. The stop signal acts as as the end-of-sentence token. When $u > 0.5$, it indicates the sentence decoder RNN should stop generating the next sentence.

\paragraph{Word decoder RNN}
After we decode the sentence topics, we can start to decode the words given the topic vector $\btau_i$.
For simplicity, we drop the subscript $i$ as this process applies to all sentences. 
We adopted the visual sentinel~\citep{lu2017knowing} that modulates the feature map $\bV$ with a sentinel vector. 
The hidden states and outputs are again modeled with LSTM, generating the posterior probability $\bp_j$ over the vocabulary with (1) the mean visual feature $\bar\bv$, (2) the topic vector $\btau$, and (3) the embedding of the previously generated word $\be_{j-1} = \bE_{\word_{j-1}}$, where $\bE \in \real^{d \times \abs{\voc}}$ is the trainable word embedding matrix. 
At training time, the next word is sampled from the probability $z_j \sim p\para{z \mid \cdot} = \para{\bp_j}_z$, or the $z$-th element of $\bp_j$.

This formulation enables the model to look at different parts on the image while having the option of ``looking away'' at a sentinel vector. Note that this hierarchical encoder-decoder CNN-RNN-RNN architecture is fully differentiable.

\subsection{Reinforcement Learning for Readability}
As \citet{rennie2017self} showed, the automatic NLG metric CIDEr~\citep{vedantam2015cider} is superior to other metrics such as BLEU~\citep{papineni2002bleu}, and ROUGE~\citep{lin2004rouge}. We consider the case of self-critical sequence training (SCST)~\citep{rennie2017self} which utilizes REINFORCE~\citep{williams1992simple} algorithm, and minimize the negative expected reward as a function of the network parameters $\theta$, as
$
    \loss_{\auto}\para{\theta} = 
    -\expect_{
        \para{u, \report} \sim 
        p_\theta\para{u, \report}
    }
    \param{
        r_{\auto}\para{\report, \report^\ast} - 
        r_{\auto}\para{\report^g, \report^\ast}
    },
$
where $p_\theta$ is the distribution over output spaces, $r_{\auto}$ is a metric evaluation function acting as a reward function that takes a sampled report $\report$ and a ground truth report $\report^\ast$. 
The baseline in SCST has been replaced with the reward obtained with testing time greedily decoded report $\report^g$.


\subsection{Novel Reward for Clinically Accurate Reinforcement Learning}
\label{sec:ccr}
One major downside with the approach outlined so far, unfortunately, is that in the clinical context, aiming for a good automatic metric such as CIDEr is not enough to correctly characterize the disease states.
Negative judgments on diseases are critical components of the reports, by which radiologist indicates that the patient might not have those diseases that were of concern and among the reasons for the examination.
\cite{li2018hybrid} indicated that a good portion of chest X-ray reports are heavily templated in patterns such as \emph{no pneumothorax or pleural effusion}; \emph{the lungs are clear}; or \emph{no focal consolidation, pneumothorax or large pleural effusion}. 
These patterns also suggest that most patients are disease-free, hence the signal of positive mentions of the disease will be sparse.

Simply optimizing the automatic LG metrics may misguide the model to mention only the disease names as opposed to correctly positively/negatively describe the disease states.
For example, if the ground truth report reads \emph{no pleural effusion}, the models would prefer the text \emph{mild pleural effusion} over unrelated text or even an empty string, which means intelligent optimization systems could game these metrics at the expense of clinical accuracy. 

We hence propose using a \emph{\oursfullupper (\oursabbr)}, which utilizes a rule-based disease mention annotator , CheXpert~\citep{irvin2019chexpert}, to optimize our generated report for clinical efficacy directly. CheXpert performs classification on 12 types of thoracic diseases or X-ray related diagnoses. The mentions for support devices are also labeled. For each label type $t$, there are four possible outcomes for the labeling: (1) positive, (2) negative, (3) uncertain, or (4) absent mention; or, $l_t\para{\report} \in \paral{\posmen, \negmen, \uncmen, \absmen}$. This outcome can be used to model the positive/negative disease state $s_t \in \paral{\posdis, \negdis}$ as $s_t \sim p_{s | l} \para{\cdot | l_t\para{\report}}$, the value of which will be discussed further later. \oursabbr is then defined, dropping the subscripts for distribution for convenience, as
\begin{small}
\begin{align}
    r_{\ourseqn}\para{\report, \report^\ast}
    =
        \sum_t r_{\ourseqn, t}\para{\report, \report^\ast}
    \equiv
        \sum_t \sum_{s \in \paral{\posdis, \negdis}}
            p\para{s | l_t\para{\report}} \cdot
            p\para{s | l_t\para{\report^\ast}},
\end{align}
\end{small}%
aiming to maximize the correlation of distribution over disease states between the generated text $\report$ and the ground truth text $\report^\ast$. Unfortunately, as the true diagnostic state $s$ of novel reports is unknown, we need to make several assumptions regarding the performance of the rule based labeler, allowing us to infer the necessary conditional probabilities $p\para{s|l}$. 

To motivate these assumptions, first note that these diseases are universally rare, or, $p\para{\posdis} \ll p\para{\negdis}$. 
Presuming the rule based labeler has any discriminative power, we can thus conclude that if the labeler assigns a negative or an absent label ($l^-$ is one of $\paral{\negmen, \absmen}$), $p\para{\posdis | l^-} < p\para{\posdis} \ll p\para{\negdis} < p\para{\negdis | l^-}.$ 
For sufficiently rare conditions, a reasonable assumption and simplification is to therefore take $p\para{\posdis | l^-} \approx 0$ and $p\para{\negdis | l^-} \approx 1$. We further assume that the rule based labeler has a very high precision, and thus $p\para{\posdis|\posmen} \approx 1$. 
However, given an uncertain mention $\uncmen$, the desired output probabilities are difficult to assess. 
As such, we define a reward-specific hyperparameter $\beta_\uncmen \equiv p\para{\posdis|\uncmen}$, which in this work we take to be $0.5$. All of these assumptions could be easily adjusted, but they perform well for us here.

We also wish to use a baseline for the reward $r_{\ourseqn}$. Instead of using a single exponential moving average (EMA) over the total reward, we apply EMA separately to each term as 
\begin{small}
\begin{align}
    \loss_{\ourseqn}\para{\theta} = 
    -\expect_{
        \para{u, \report} \sim 
        p_\theta\para{u, \report}
    }
    \param{
        \sum_t
        r_{\ourseqn, t}\para{\report, \report^\ast} -
        \bar{r}_{\ourseqn, t}
    },
\end{align}
\end{small}%
where $\bar{r}_{\ourseqn, t}$ is an EMA over $r_{\ourseqn, t}$ updated as $\bar{r}_{\ourseqn, t} \leftarrow \gamma \bar{r}_{\ourseqn, t} + \para{1 - \gamma} r_{\ourseqn, t}\para{\report, \report^\ast}$.

We wish to pursue both semantic alignment and clinical coherence with the ground truth report, and thus we combine the above rewards for reinforcement learning on our neural network in a weighted fashion.
Specifically, $\loss\para{\theta} = \loss_{\auto}\para{\theta} + \lambda \loss_{\ourseqn}\para{\theta}$, where $\lambda$ controls the relative importance.

Hence the derivative of the combined loss with respect to $\theta$ is thus
\begin{small}
\begin{align}
    \nabla_\theta \loss\para{\theta} 
    =
        -\expect_{
            \para{u, \report} \sim 
            p_\theta\para{u, \report}
        }
        \param{
            \param{
                r_{\auto}\para{\report, \report^\ast} 
                + \lambda r_{\ourseqn}\para{\report, \report^\ast}
            }
            \nabla_\theta\sum_i \para{
                \log u_i + \sum_j \log \para{\bp_{ij}}_{\word_{ij}}
            }
        },
\end{align}
\end{small}
where $\bp_{ij}$ is the probability over vocabulary. We can approximate the above gradient with Monte-Carlo samples from $p_\theta$ and average gradients across training examples in the batch.
\section{Experiments}

\subsection{Datasets}
In this work, we use two chest X-ray/report datasets: MIMIC-CXR~\citep{johnson2019mimic} and Open-I~\citep{demner2015preparing}. 

MIMIC-CXR is the largest radiology dataset to date and consists of $473,057$ chest X-ray images and $206,563$ reports from $63,478$ patients\footnote{This work used an alpha version of MIMIC-CXR instead of the publicly released version where the images are more standardized and the split into official train/test sets.}. Among these images, $240,780$ are of anteroposterior (AP), $101,379$ are of posteroanterior (PA), and $116,023$ are of lateral (LL) views. 
Furthermore, we eliminate duplicated radiograph images with adjusted brightness level or contrast as they are commonly produced for clinical needs, after which we are left with $327,281$ images and $141,783$ reports.
The radiological reports are parsed into sections, among which we extract the \emph{findings} section. We then apply tokenization and keep tokens with at least $5$ occurrences in the corpus, resulting in $5,571$ tokens in total.

Open-I is a public radiography dataset collected by Indiana University with $7,471$ chest X-ray images and $3,955$ reports. The reports are in XML format and include pre-parsed sections. We then exclude the entries without the \emph{findings} section and are left with $6,471$ images and $3,336$ reports. Tokenization is done similarly, but due to the relatively small size of the corpus, we keep tokens with $3$ or more occurrences, ending up with $948$ tokens.

Both datasets are partitioned by patients into a train/validation/test ratio of 7/1/2 so that there is no patient overlap between sets. Words that are excluded were replaced by an ``unknown'' token, and the word embeddings are pretrained separately for each dataset. 

\subsection{Evaluation Metrics}
To compare with other models including prior state-of-the-art and baselines, we adopt several different metrics that focus on different aspects ranging from a natural language perspective to clinical adequacy. 

Automatic LG metrics such as CIDEr-D~\citep{vedantam2015cider}, ROUGE-L~\citep{lin2004rouge}, and BLEU~\citep{papineni2002bleu} measure the statistical relation between two text sequences. One concern with such statistical measures is that with a limited scope from the $n$-grams ($n$ up to $4$) we are unable to capture disease states, as negations are common in the medical corpus and oftentimes the negation cue words and disease words can be far apart in a sentence. 
As such, we also include medical abnormality detection as a metric. Specifically, we compare the CheXpert~\citep{irvin2019chexpert} labeled annotations between the generated report and the ground truth report on $14$ different categories related to thoracic diseases and support devices\footnote{We decide not to include NegBio~\citep{peng2018negbio}, a previous state-of-the-art disease labeling system, due to its significant performance gap with CheXpert as reported \citet{irvin2019chexpert} and \citet{johnson2019mimic}}. We evaluate the accuracy, precision, and recall for all models.

\subsection{Models}
We compare our methods with state-of-the-art image captioning and medical report generation models as well as some simple baseline models: (a) \emph{1-NN}, in which we query in the image embedding space for the closest neighbor in the train set using a test image. The corresponding report of the neighbor is used as the output for this test image;
(b) \emph{Show and Tell} (S\&T)~\citep{vinyals2015show}; (c) \emph{Show, Attend, and Tell} (SA\&T)~\citep{xu2015show}; and (d) \emph{TieNet}~\citep{wang2018tienet}.
To allow comparable results in all models, we slightly modify previous models to also accept the view position embedding which encodes AP/PA/LL as a one-hot vector to utilize the extra information available at image acquisition. This includes \emph{Show and Tell}, \emph{Show, Attend, and Tell}, and our re-implementation of \emph{TieNet}, which is detailed in Appendix~\ref{sec:TieNet} because the authors did not release their code.

We observed our model to sometimes repeat the findings multiple times. We apply post-hoc processing where we remove exact duplicate sentences in the generated reports. This proves to improve the readability but interestingly slightly degrades NLG metrics. 

Additionally, we perform several ablation studies to inspect the contribution of various components of our model. In particular, we assess
\begin{description}
    \item[Ours ($\auto$)] Use $r_{\auto}$ only for reinforced learning, as often is the case with the prior state-of-the-art. 
    \item[Ours ($\ourseqn$)] Use $r_{\ourseqn}$ only and do not care about aligning the natural language metrics.
    \item[Ours (full)] Considers both rewards as formulated in Section~\ref{sec:ccr}.
\end{description}

In order to provide some context to the metric scores, we also trained an unsupervised RNN language model which generates free text without conditioning on input radiograph images, which we denote as \emph{Noise-RNN}. All recurrent models, including prior works and our models, use beam search with a beam size of $4$.
\section{Results \& Discussion}
\subsection{Quantitative Results}
\paragraph{Natural Language Metrics}
\begin{table}[!t]
\setlength{\tabcolsep}{0.4em}
\centering
\footnotesize

\begin{tabular}{C{0.3cm} |	L{2.2cm} |	C{1.35cm}	C{1.35cm}	C{1.35cm}	C{1.35cm}	C{1.35cm}	C{1.35cm} |	C{1.35cm}}
\toprule
 	& \multirow{2}{*}{\bf Model}	& \multicolumn{6}{c|}{\bf Natural Language}	& \bf Clinical\\ \cline{3-9}
 	&  	& CIDEr	& ROUGE	& BLEU-1	& BLEU-2	& BLEU-3	& BLEU-4	& Accuracy\\ \cline{1-9}
\multirow{9}{*}{\rotatebox[origin=c]{90}{\bf MIMIC-CXR}}	& \emph{Major Class}	& -	& -	& -	& -	& -	& -	& 0.828\\ 
 	& Noise-RNN	& 0.716	& 0.272	& 0.269	& 0.172	& 0.113	& 0.074	& 0.803\\ \cline{2-9}
 	& 1-NN	& 0.755	& 0.244	& 0.305	& 0.171	& 0.098	& 0.057	& 0.818\\ 
 	& S\&T	& 0.886	& 0.300	& 0.307	& 0.201	& 0.137	& 0.093	& 0.837\\ 
 	& SA\&T	& 0.967	& 0.288	& 0.318	& 0.205	& 0.137	& 0.093	& 0.849\\ 
 	& TieNet	& 1.004	& 0.296	& 0.332	& 0.212	& 0.142	& 0.095	& 0.848\\ \cline{2-9}
 	& Ours ($\auto$)	& \bf 1.153	& \bf 0.307	& \bf 0.352	& \bf 0.223	& \bf 0.153	& \bf 0.104	& 0.834\\ 
 	& Ours ($\ourseqn$)	& 0.956	& 0.284	& 0.294	& 0.190	& 0.134	& 0.094	& \bf 0.868\\ 
 	& Ours (full)	& 1.046	& \bf 0.306	& 0.313	& 0.206	& 0.146	& \bf 0.103	& \bf 0.867\\ \hhline{=|=|======|=}
\multirow{9}{*}{\rotatebox[origin=c]{90}{\bf Open-I}}	& \emph{Major Class}	& -	& -	& -	& -	& -	& -	& 0.911\\ 
 	& Noise-RNN	& 0.747	& 0.291	& 0.233	& 0.130	& 0.087	& 0.061	& 0.914\\ \cline{2-9}
 	& 1-NN	& 0.728	& 0.201	& 0.232	& 0.116	& 0.051	& 0.018	&  0.911\\ 
 	& S\&T	& 0.926	& 0.306	& 0.265	& 0.157	& 0.105	& 0.073	& 0.915\\ 
 	& SA\&T	& 1.276	& 0.313	& 0.328	& 0.195	& 0.123	& 0.080	&  0.908\\ 
 	& TieNet	& 1.334	& 0.311	& 0.330	& 0.194	& 0.124	& 0.081	& 0.902\\ \cline{2-9}
 	& Ours ($\auto$)	& \bf 1.490	& \bf 0.359	& \bf 0.369	& \bf 0.246	& \bf 0.171	& \bf 0.115	& 0.916\\ 
 	& Ours ($\ourseqn$)	& 0.707	& 0.244	& 0.162	& 0.084	& 0.055	& 0.036	& \bf 0.917\\ 
 	& Ours (full)	& 1.424	& 0.354	& 0.359	& 0.237	& 0.164	& \bf 0.113	& \bf 0.918\\ 
\bottomrule
\end{tabular}

\caption{\small \textbf{Automatic Evaluation Scores.} The table is divided into natural language metrics and clinical finding accuracy scores. BLEU-$n$ counts up $n$-gram for evaluation, and accuracy is the averaged macro accuracy across all clinical findings. \emph{Major class} always predicts negative findings.}
\label{tbl:q-1}
\vspace{-20pt}

\end{table}

In Table~\ref{tbl:q-1} we show the automatic evaluation scores for baseline models, prior works, and variants of our models on the aforementioned test sets. 
Ours ($\auto$), that solely optimizes CIDEr score, achieves superior performance in terms of natural language metrics, but its clinical meaningfulness is not significantly above the \emph{major class} in which we predict all patients to be disease-free. 
This phenomenon is common among all other models that do not consider the clinical alignment between the ground truth and the generated reports.
On the other hand, in our full model, if we consider both natural language and clinical coherence, we can achieve the highest clinical disease annotation accuracy while still retaining decently high NLG metrics. 

We also conducted the ablation study with the model variant Ours ($\ourseqn$), where we use reinforcement learning on only the clinical accuracy. It is clear that we are unable to achieve higher clinical coherence, though readability might be sacrificed. We thus conclude that a combination of both NLG metrics and a clinically sensible objective is crucial in training a useful model in clinical practice.

One thing to note is that although \emph{Noise-RNN} is not dependent on the image, its NLG metrics, especially ROUGE, are not far off from models learned with supervision. We also note that MIMIC-CXR is better for training such an encoder-decoder model not just for its larger volume of data, but also due to its higher proportion of positive disease annotations at $16.7\%$ while Open-I only has $5.4\%$. This discrepancy leads to a 156 times difference in the number of images from diseased patients.

\paragraph{Clinical Efficacy Metrics}
\begin{table}[!t]
\setlength{\tabcolsep}{0em}
\centering
\footnotesize

\begin{tabular}{C{0.5cm} |	L{0.15cm}	L{4.15cm}	C{1.2cm} |	C{1.32cm}	C{1.32cm}	C{1.32cm}	C{1.32cm} | C{1.32cm}	C{1.32cm}	C{1.32cm}}
\toprule
  	&  	& Label	& Count	& 1-NN	& S\&T	& SA\&T	& TieNet	& \makecell{Ours\\($\auto$)}	& \makecell{Ours\\($\ourseqn$)}	& \makecell{Ours\\(full)}\\ \hline
 \multirow{19}{*}{\rotatebox[origin=c]{90}{\bf MIMIC-CXR}}	&  	& Total	& 69031	& -	& -	& -	& -	& -	& -	& -\\ \cline{3-11}
  	&  	& No Finding	& 15677	& 0.432	& 0.299	& 0.349	& 0.339	& 0.339	& \bf 0.491	& 0.405\\
  	&  	& Enlarged Cardiomediastinum	& 6064	& 0.123	& 0.134	& 0.163	& 0.179	& 0.180	& \bf 0.202	& 0.167\\
  	&  	& Cardiomegaly	& 19065	& 0.440	& 0.535	& 0.438	& 0.464	& 0.000	& 0.678	& \bf 0.704\\
  	&  	& Lung Lesion	& 2447	& 0.064	& \bf 0.333	& 0.223	& 0.000	& 0.000	& 0.000	& 0.000\\
  	&  	& Airspace Opacity	& 21972	& 0.432	& 0.607	& 0.592	& 0.571	& 0.453	& \bf 0.640	& 0.460\\
  	&  	& Edema	& 6594	& 0.265	& 0.331	& 0.244	& \bf 0.405	& 0.266	& 0.280	& 0.000\\
  	&  	& Consolidation	& 2384	& 0.076	& 0.013	& \bf 0.180	& 0.151	& 0.089	& 0.037	& 0.000\\
  	&  	& Pneumonia	& 3068	& 0.065	& 0.106	& 0.091	& 0.082	& 0.075	& 0.000	& \bf 0.400\\
  	&  	& Atelectasis	& 16161	& 0.374	& 0.490	& 0.496	& 0.470	& 0.385	& 0.476	& \bf 0.521\\
  	&  	& Pneumothorax	& 2636	& 0.079	& 0.034	& 0.095	& 0.081	& 0.081	& 0.039	& \bf 0.098\\
  	&  	& Pleural Effusion	& 15283	& 0.534	& 0.550	& 0.545	& \bf 0.735	& 0.487	& 0.683	& 0.689\\
  	&  	& Pleural Other	& 1285	& \bf 0.039	& 0.000	& 0.103	& 0.000	& 0.000	& 0.000	& 0.000\\
  	&  	& Fracture	& 2617	& \bf 0.059	& 0.000	& 0.000	& 0.000	& 0.000	& 0.000	& 0.000\\
  	&  	& Support Devices	& 22227	& 0.534	& 0.823	& 0.847	& 0.827	& 0.794	& 0.849	& \bf 0.880\\ \cline{3-11}
  	&  	& Precision (macro)	&  	& 0.253	& 0.304	& \bf 0.312	& 0.307	& 0.225	& \bf 0.313	& 0.309\\
  	&  	& Precision (micro)	&  	& 0.383	& 0.414	& 0.430	& 0.473	& 0.419	& \bf 0.634	& 0.586\\ \hhline{~|===|=======}
  	&  	& Recall (macro)	&  	& \bf 0.265	& 0.173	& 0.232	& 0.220	& 0.209	& 0.126	& 0.134\\
  	&  	& Recall (micro)	&  	& \bf 0.400	& 0.276	& 0.367	& 0.355	& 0.360	& 0.227	& 0.237\\
\bottomrule
\end{tabular}

\caption{\small \textbf{Clinical Finding Scores}. The precision scores for each of the labels are listed and aggregated into the overall precision scores. Recall scores are shown in the last two rows. Macro denotes averaging the numbers in the table directly and micro accounts for class prevalence.}
\label{tbl:q-2}
\vspace{-20pt}

\end{table}

In Table~\ref{tbl:q-2} we can compare the labels annotated by CheXpert calculated over all test set generated reports.
Note that the labeling process generates discrete binary label as opposed to predicting continuous probabilities, and as such we are unable to obtain discriminative metrics such as the Area Under the Receiver Operator Characteristic (AUROC) or the Area Under the Precision-Recall Curve (AUPRC). Precision-wise, Ours ($\ourseqn$) achieves the highest overall scores including macro-average and micro-average. 
The runner-up is Ours (full) model, which additionally considers language fluency. Note that the macro- metrics can be quite noisy as the per-class metric can be dependent on just a few examples. Many entries in the table are zeros, as they never yield positive predictions and we regard them as zeros to penalize such behavior.
Regarding the recall metric, we are able to see a substantial drop in Ours ($\ourseqn$) and Ours (full) as a result of optimizing for accuracy. Accuracy is closely associated with precision but overpursuing it might harm in terms of recall.
It is worthwhile to notice that the nearest neighbor \emph{1-NN} has the highest recall, and this is no surprise since as shown before~\citep{gltr}, generated sequences tend to follow the statistics and favor common words too much. 
Rare combinations of tokens in the corpus can be easily neglected, resulting in predictions of mostly major classes.

\subsection{Qualitative Results}
\paragraph{Evaluation of Generated Reports}
\begin{table}[!t]
\setlength{\tabcolsep}{0.4em}
\centering
\tiny
\sffamily

\begin{tabular}{L{2.0cm}	L{4.7cm}	L{4.0cm}	L{3.6cm}}

  	& \multicolumn{1}{c}{\bf Ground Truth}	& \multicolumn{1}{c}{\bf TieNet}	& \multicolumn{1}{c}{\bf Ours (full)}\\ \cline{2-4}
 \includegraphics[width=\linewidth]{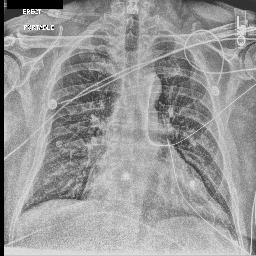}	& cardiomegaly is moderate. bibasilar atelectasis is mild. there is no pneumothorax. a lower cervical spinal fusion is partially visualized. healed right rib fractures are incidentally noted.	& ap portable upright view of the chest. there is no focal consolidation, effusion, or pneumothorax. the cardiomediastinal silhouette is normal. imaged osseous structures are intact.	& pa and lateral views of the chest. there is mild enlargement of the cardiac silhouette. there is no pleural effusion or pneumothorax. there is no acute osseous abnormalities.\\ \cline{2-4}
 \includegraphics[width=\linewidth]{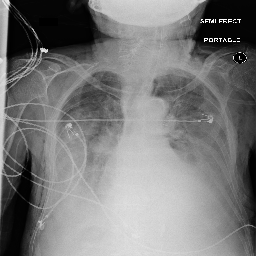}	& as compared to the previous radiograph, the monitoring and support devices are unchanged. unchanged bilateral pleural effusions, with a tendency to increase, and resultant areas of atelectasis. the air collection in the bilateral soft tissues is slightly decreased. unchanged right picc line. no definite evidence of pneumothorax.	& as compared to the previous radiograph, the patient has received a nasogastric tube. the course of the tube is unremarkable, the tip of the tube projects over the middle parts of the stomach. there is no evidence of complication, notably no pneumothorax. the other monitoring and support devices are constant. constant appearance of the cardiac silhouette and of the lung parenchyma.	& as compared to the previous radiograph, there is no relevant change. tracheostomy tube is in place. there is a layering pleural effusions. NAME bilateral pleural effusion and compressive atelectasis at the right base. there is no pneumothorax.\\ \cline{2-4}
 \includegraphics[width=\linewidth]{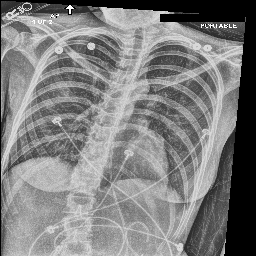}	& the course of the dobbhoff feeding catheter is unremarkable, and its tip is seen projecting over the stomach. there is no evidence of complications, specifically no pneumothorax. as compared to the prior radiograph dated DATE, there has been no other significant interval change.	& ap portable upright view of the chest. overlying ekg leads are present. there is no focal consolidation, effusion, or pneumothorax. the cardiomediastinal silhouette is normal. imaged osseous structures are intact.	& as compared to the previous radiograph, there is no relevant change. the endotracheal tube terminates approximately 3 cm above the NAME. the endotracheal tube extends into the stomach. there is no evidence of complications, notably no pneumothorax. there is no pleural effusion or pneumothorax.\\ \cline{2-4}
 \includegraphics[width=\linewidth]{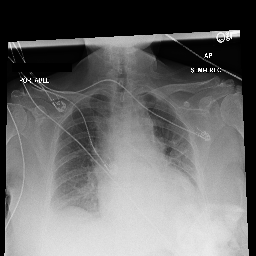}	& interval placement of a left basilar pigtail chest tube with improving aeration in the left mid to lower lung and near complete resolution of the pleural effusion. there are residual patchy opacities within the left mid and lower lung as well as at the right base favoring resolving atelectasis. no pneumothorax is appreciated on this semi upright study. heart remains stably enlarged. mediastinal contours are stably widened, although this NAME be related to portable technique and positioning. this can be better evaluated on followup imaging. no pulmonary edema.	& as compared to the previous radiograph, the patient has been extubated. the nasogastric tube is in unchanged position. the lung volumes remain low. moderate cardiomegaly with minimal fluid overload but no overt pulmonary edema. no larger pleural effusions. no pneumonia.	& ap upright and lateral views of the chest. there is moderate cardiomegaly. there is no pleural effusion or pneumothorax. there is no acute osseous abnormalities.\\

\end{tabular}

\rmfamily
\caption{\small \textbf{Sample images along with ground truth and generated reports}. Note that upper case tokens are results of anonymization.}
\label{tbl:samples}
\vspace{-15pt}

\end{table}

Table~\ref{tbl:samples} demonstrates the qualitative results of our full model. 
In general, our models are able to generate descriptions that align with the logical flow of reports written by radiologists, which start from general information (such as views, previous comparison), positive, then negative findings, with the order of lung, heart, pleura, and others.
TieNet also generates report descriptions with such logical flow but in slightly different orders.
For the negative findings cases, both our model and TieNet do well on generating reasonable descriptions without significant errors.
Regarding the cases with positive findings, TieNet and our full model both cannot identify all radiological findings. 
Our full model is able to identify the major finding in each demonstrated case. 
For example, cardiomegaly in the first case, pleural effusion, and atelectasis in the second case. 

A formerly practicing clinician co-author reviewed a larger subset of our generated reports manually. They drew several conclusions. First, our full model tends to generate sentences related to pleural effusion, atelectasis, and cardiomegaly correctly---which is aligned with the clinical finding scores in Table~\ref{tbl:q-2}.
TieNet instead misses some positive findings in such cases.
Second, there are significant issues in \emph{all} generated reports, regardless of the source model, which include the description of supportive lines and tubes, as well as lung lesions.
For example, TieNet is prone to generate nasogastric tube mentions while our model tends to mention tracheostomy or endotracheal tube, and yet both models have difficulty identifying some specific lines such as chest tube or PICC line. Similarly, both systems do not generate the sentence with positive lung parenchymal findings correctly. 

From this (small) sample, we are unable to draw a conclusion whether our model or TieNet truly outperforms the other since both present with significant issues and each has strengths the other lacks. Critically, neither of them can describe the majority of the findings in the chest radiograph well, especially for positive cases, even if the quantitative metrics demonstrate the reasonable performance of the models. This illustrates that \emph{significant} progress is still needed in this domain, perhaps building on the directions we explore here before these techniques could be deployed in a clinical environment.

\paragraph{Learning Meaningful Attention Maps}
\begin{figure}[!t]
\centering
\includegraphics[width=\linewidth]{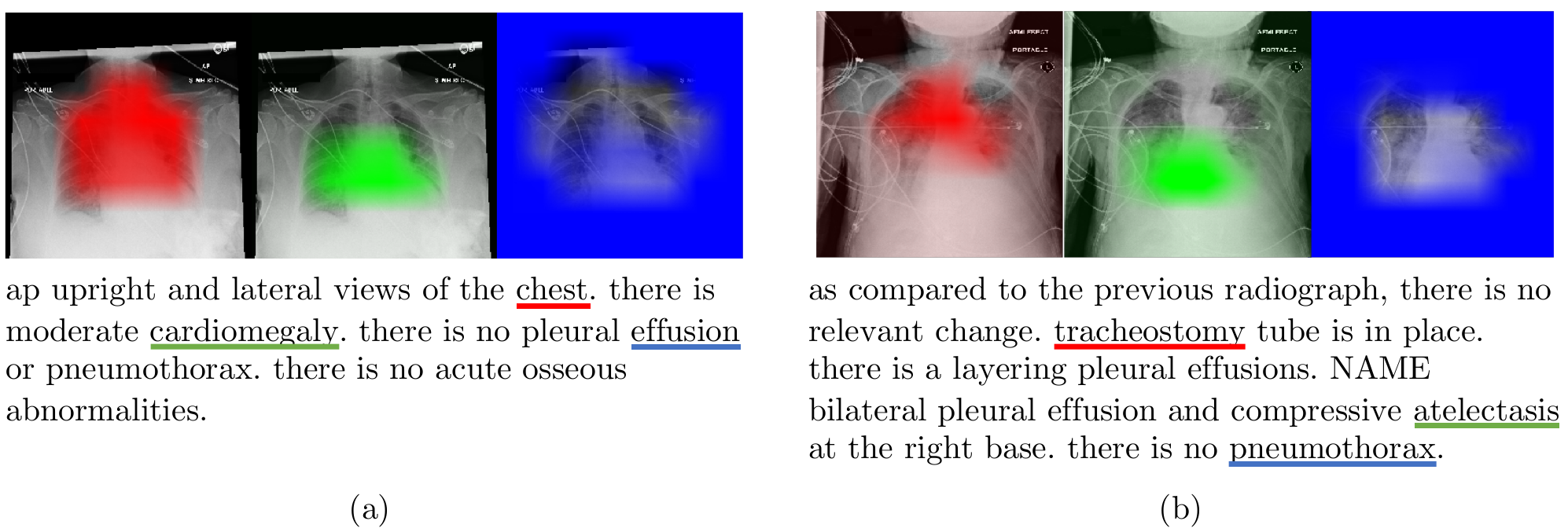}
\caption{\small \textbf{Visualization of the generated report and image attention maps.} Different words are underlined with its corresponding attention map shown in the same color. Best viewed in color.}
\label{fig:attention-map}
\vspace{-20pt}
\end{figure}
Attention maps have been a useful tool in visualizing what a neural network is attending to, as demonstrated by \citet{rajpurkar2017chexnet}.
Figure~\ref{fig:attention-map} shows the intermediate attention maps for each word when it is being generated. As we can observe, the model is able to roughly capture the location of the indicated disease or parts, but we also find, interestingly, that the attention map tends to be the complement of the actual region of interest when the disease keywords follow a negation cue word. This might indicate that the model is actively looking at the rest of the image to ensure it does not miss any possible symptoms exhibited before asserting disease-free states. This behavior has not been widely discussed before, partially because attention maps for negations are not the primary focus of typical image captioning tasks, and most attention mechanisms employed in a clinical context were on classification tasks where they also do not specifically focus on negations.
\section{Conclusion}
\subsection{Limitations \& Future Work}
Our work has several notable limitations and opportunities for future work. First and foremost, the post-processing step required to remove repeated sentences is an ugly necessity, and we endeavor to remove it in future iterations of this work. Promising techniques exist in NLG for the inclusion of greater diversity, which warrant further investigation here.

Secondly, our model operates using images in isolation, without consideration of whether these images are part of a series of ordered radiographs for a single patient, which might be summarized together. Using all available information has the potential to improve the quality of the generated reports, and should definitely be investigated further.

Lastly, we note that though our model yields very strong performance for CheXpert \emph{precision}, its recall is much worse. Recall versus precision is favored to different degrees in differing clinical contexts. For example, for screening purpose, recall (sensitivity) is an ideal metric since the healthy cases usually won't give positive findings. However, precision (positive predictive value) is much more critical for validating the clinical impression, which is common in an ICU setting where patients receive a radiological study on the basis of strong clinical suspicion. We believe that our system's poor recall is a direct result of the setup of our RL models and the CCR reward, which optimizes for accuracy and inherently boosts precision. 
It is the choice of optimization objectives that lead to the results. 
Depending on the actual clinical applications, we may, in turn, optimize \emph{Recall at Fixed Precision} (R@P) or $F_\beta$ score via methods described by \citet{eban2016scalable}.


\subsection{Reflections on Trends in the Field}
In the course of this work, we also encounter several other larger points which are present not only in our study but also in many related studies in this domain and warrant further thought by the community.

\paragraph{System Generalizability}
CheXpert used in our models is rule-based, which is harder to generalize to other datasets and to identify the implicit features inside the language patterns. 
CheXpert is also specialized in English and would require considerable work to re-code its rules for other natural languages.
A more universal approach for subsequent research may use a learning-based approach for labeling to improve generalizability and extend to corpora in different languages;
for example, PadChest in Spanish.

\paragraph{Be Careful What You Wish For}
NLG metrics are known to be only limited substitutes for a true assessment of readability~\citep{kilickaya2016re,liu2016not}. For radiology reports more specifically, this problem is even more profound, as prior works often use ``readability'' as a proxy for clinical efficacy.
Additionally, we note that these NLG evaluation metrics are easily susceptible to gaming. In our results, our post-processing step of removing exact duplicates actually \emph{worsens} our CIDEr score, which is the opposite of what should be desired for an NLG evaluation metric. 
Even if our proposed clinical coherence aims at resolving the unwanted misalignment between NLG and real practice, we are not able to obviously judge whether our system is better despite its performance on paper.
This fact is especially troubling given the increasing trend of using reinforcement learning (RL) to directly optimize objectives, as has been done in prior work~\citep{li2018hybrid} and as we do here. 
Though RL can offer marked improvements in these automatic metrics, which are currently the best the field can do, how well it translates to the real clinical efficacy is unclear. 
The careful design of improved evaluation metrics, specifically for radiology report generation, should be a prime focus for the field going forward.


\subsection{Conclusion}
In this work, we develop a chest X-Ray radiology report generation system which hierarchically generates topics from images, then words from topics.
This structure gives the model the ability to use largely templated sentences (through the generation of similar topic vectors) while preserving its freedom to generate diverse text. The final system is also optimized with reinforcement learning for both readability (via CIDEr) and clinical correctness (via the novel Clinically Coherent Reward). Our system outperforms a variety of compelling baseline methods across readability and clinical efficacy metrics on both MIMIC-CXR and Open-I datasets.

\section*{Acknowledgements}
Dr. Marzyeh Ghassemi is partially funded by a CIFAR AI Chair at the Vector Institute, and an NSERC Discovery Grant.

Matthew McDermott is funded in part by National Institutes of Health: National Institutes of Mental Health grant P50-MH106933 as well as a Mitacs Globalink Research Award.

\bibliography{main}
\appendix

\newpage
\section{Implementation Details}
\label{sec:implementation}

We briefly describe the details of our implementation in this section.

\paragraph{Encoder}
The image encoder CNN takes an input image of size $256 \times 256 \times 3$. The last layer before global pooling in a DenseNet-121 are extracted, which has a dimension of $8 \times 8 \times 1024$, and thus $K = 64$ and $d_\phi = 1024$. Densenet-121~\citep{iandola2014densenet} has been shown to be state-of-the-art in the context of classification for clinical images. The image features are then projected to $d = 256$ dimensions with a dropout of $p=0.5$. 

Since typically in the X-ray image acquisition we are provided with the view position indicating the posture of the patient related to the machine, we conveniently pass this into the model as well. Indicated by a one-hot vector, the view position embedding is concatenated with the image embedding to form an input to the later decoders.

\paragraph{Decoder}
As previously mentioned, the input image embedding to the LSTM has a dimension of $256$, and it is the same for word embeddings and hidden layer sizes.
The word embedding matrix is pretrained with Gensim~\citep{rehurek2010software} in an unsupervised manner. 

\paragraph{Training Details}  
We implement our model on PyTorch~\citep{paszke2017automatic} and train on 4 GeForce GTX TITAN X GPUs. 
All models are first trained with cross-entropy loss with the Adam~\citep{kingma2014adam} optimizer using an initial learning rate of $10^{-3}$ and a batch size of $64$ for $64$ epochs. Other than the weights stated above, the models are initialized randomly.
Learning rates are annealed by $0.5$ every $16$ epochs and we increase the probability of feeding back a sample from the posterior $\bp$ by $0.05$ every $16$ epochs. After this bootstrapping stage, we start training with REINFORCE for another $64$ epochs.
The initial learning rate for the second stage is $10^{-5}$ and is annealed on the same schedule.

Indicated by \cite{rennie2017self}, we adopt CIDEr-D~\citep{vedantam2015cider} metric as the reward module used in $r_{\auto}$.
For the baseline for \oursabbr, we choose a EMA momentum $\gamma = 0.95$. A weighting factor $\lambda = 10$ has been chosen to balance the scales of the rewards for our full model.

\newpage
\section{TieNet Re-implementation}
\label{sec:TieNet}

Since the implementation for TieNet~\citep{wang2018tienet} is not released, we re-implement it with the descriptions provided by the original authors. The re-implementation details are described in this section.

\paragraph{Overview}
TieNet stands for \emph{Text-Image Embedding Network}. It consists of three main components: image encoder, sentence decoder with \emph{Attention Network}, and \emph{Joint Learning Network}. It computes a global attention encoded text embedding using hidden states from a sentence decoder and saliency weighted global average pooling using attention maps from the attention network. The two global representations are combined as an input to the joint learning network. Finally, it outputs the multi-label classification of thoracic diseases. The end products are automatic report generation for medical images and classification of thoracic diseases. 

\paragraph{Encoder}
An image of size $256 \times 256 \times 3$ is taken by the image encoder CNN as an input. The last two layers of ResNet-101~\citep{he2016deep} are removed since we are not classifying the image. The final encoding produced has a size of $14 \times 14 \times 2048$. We also fine-tune convolutional blocks \texttt{conv2} through \texttt{conv4} of our image encoder during training time.

\paragraph{Decoder}
We also include the view position information by concatenating the view position embedding with the image embedding to form input. The view position embedding is indicated by a one-hot vector. At each decoding step, the encoded image and the previous hidden state with a dropout of $p=0.5$ is used to generate weights for each pixel in the attention network. The previously generated word and the output from the attention network are fed to the LSTM decoder to generate the next word.

\paragraph{Joint Learning Network}
TieNet proposed an additional component to automatically classify and report thoracic diseases. The joint learning network takes hidden states and attention maps from the decoder and computes global representations for report and images, then combines the result as the input to a fully connected layer to output disease labels. 

In the original paper, $r$ indicates the number of attention heads, which we set as $r=5$; $s$ is the hidden size for attention generation, which we set as $s=2000$. 
One key difference from the original work is that we are classifying the joint embeddings into CheXpert~\citep{irvin2019chexpert} annotated labels, and hence we have the class count $M=14$.
The disease classification cross-entropy loss $L_C$ and the teacher-forcing report generation loss $L_R$ are combined as $L_{\textrm{overall}} = \alpha L_C + (1-\alpha)L_R$, in which $L_{\textrm{overall}}$ is the loss for which the network optimizes. However, the value $\alpha$ was not disclosed in the original work and we use $\alpha=0.85$.

\paragraph{Training}
We implement TieNet on PyTorch~\citep{paszke2017automatic} and train on $4$ GeForce GTX TITAN X GPUs. The decoder is trained with cross-entropy loss with the Adam~\citep{kingma2014adam} optimizer using an initial learning rate of $10^{-3}$ and a mini-batch size of $32$ for $64$ epochs. Learning rate for the decoder is decayed by a factor of $0.2$ if there is no improvement of BLEU~\citep{papineni2002bleu} score on the development set in $8$ consecutive epochs. The joint learning network is trained with sigmoid binary cross-entropy loss with the Adam~\citep{kingma2014adam} optimizer using a constant learning rate of $10^{-3}$.

\paragraph{Result}
Since we are not able to access the original implementation of TieNet and we additionally inject view position information to the model, we might have small variations in result between the original paper and our re-implementation. We only compare the report generation part of TieNet to our model. 
\end{document}